\def\year{2019}\relax
\documentclass[letterpaper]{article} %DO NOT CHANGE THIS
\usepackage{aaai19}  %Required
\usepackage{times}  %Required
\usepackage{helvet}  %Required
\usepackage{courier}  %Required
\usepackage{url}  %Required
\usepackage{graphicx}  %Required
\usepackage{amsfonts}
\usepackage{color,xcolor}%%%%%%%%%%%%%%%%%%%%%
\begin{document}
\title{PCGAN: Partition-Controlled Human Image Generation}
\author{Dong Liang$^*$,
Rui Wang\thanks{The first two authors contributed equally to this work.}\thanks{Corresponding author.},
Xiaowei Tian,
Cong Zou\\
SKLOIS, Institute of Information Engineering, Chinese Academy of Sciences\\
School of Cyber Security, University of Chinese Academy of Sciences\\
{\tt\small \{liangdong, wangrui, tianxiaowei, zoucong\}@iie.ac.cn}
}

\maketitle
\begin{abstract}
Human image generation is a very challenging task since it is affected by many factors. Many human image generation methods focus on generating human images conditioned on a given pose, while the generated backgrounds are often blurred. In this paper, we propose a novel Partition-Controlled GAN to generate human images according to target pose and background. Firstly, human poses in the given images are extracted, and foreground/background are partitioned for further use. Secondly, we extract and fuse appearance features, pose features and background features to generate the desired images. Experiments on Market-1501 and DeepFashion datasets show that our model not only generates realistic human images but also produce the human pose and background as we want. Extensive experiments on COCO and LIP datasets indicate the potential of our method.
\end{abstract}

\section{Introduction}
\label{Introduction}
Photo-realistic image editing via computer programs is an attracting idea in the computer vision field. Human image editing or generation is one of the most challenging topics. This is because our visual system is too familiar with human figures and surrounding backgrounds, which leads to a low tolerance rate for flaws in the generated images.

Recent researches on similar image generation tasks have produced fruitful results. Works based on Generative Adversarial Network (GAN) \cite{NIPS2014_5423} achieve the task by adapting image translation between two domains. By training the GAN model on two image domains of face, AttGAN \cite{he2017arbitrary} can change the foreground face in the source image according to the target face attributes. In a more general sense, the CycleGAN \cite{CycleGAN2017} translates domains with more complex changes, such as horse and zebra, photo and painting. However, these methods generate images with appearance changes only or sometimes with tiny shape changes according to the differences between domains rather than instances. The human image generation task requires image translations with non-trivial movement and even challenging changes with only semantic similarity between images.

In this paper, we aim to generate a human image using the appearance of a person in a source image, and the pose as well as background information in the target image. Specifically, we turn this task into image generation by synthesizing two parts: (1) the appearance of the foreground in the source image and (2) the specific pose and background in the target image. This appearance of person is mainly related to the color and style of the clothes. While, in order to fit the pose and background of the target image, the non-rigid transformation should be applied to the source image.

\begin{figure}
  \centering
  \includegraphics[width=\linewidth]{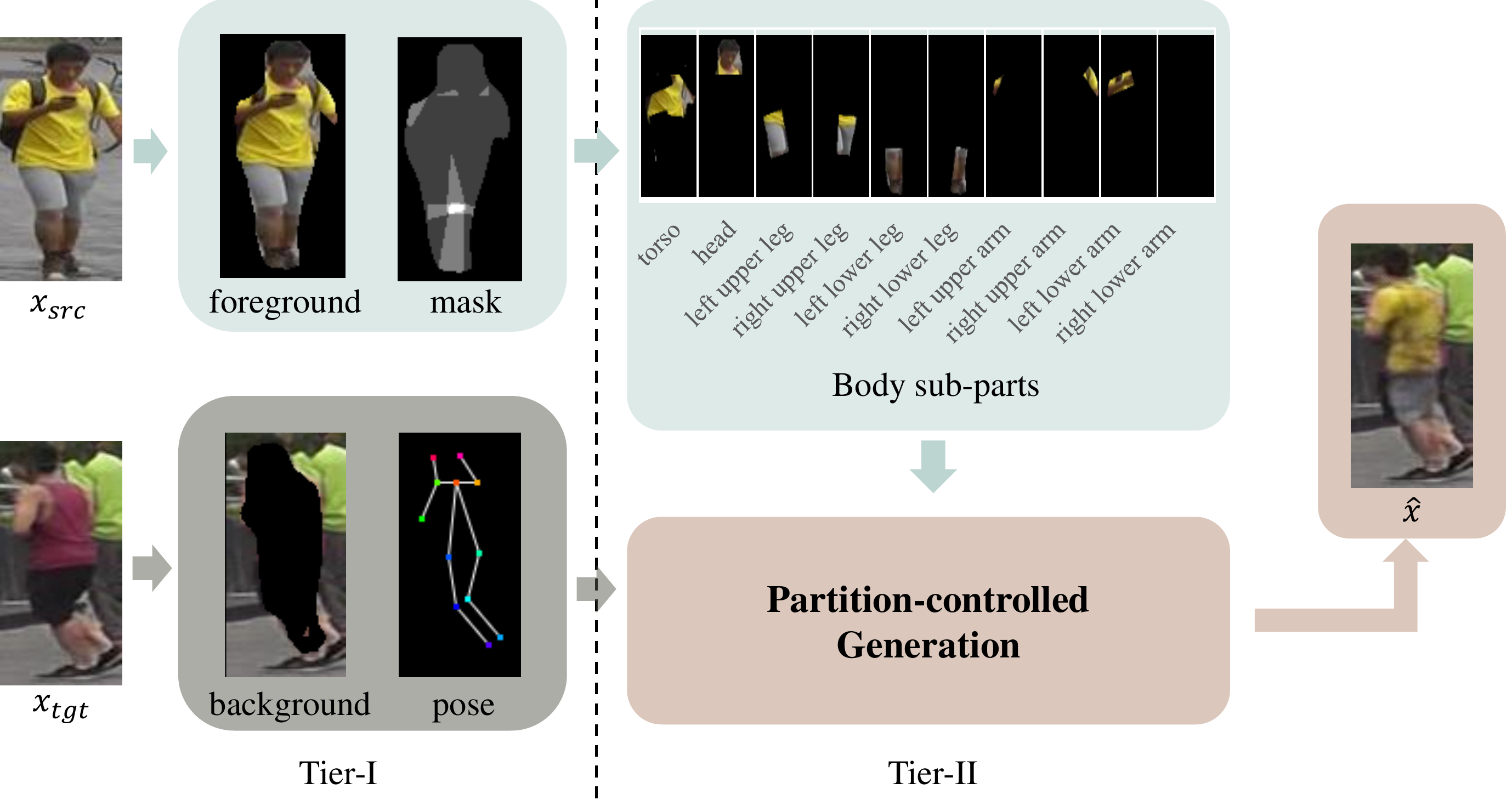}
  \caption{The partition-controlled human image generation (test phase). The proposed algorithm is two-tier: (1) Partition the foreground human body and the background and extracting the landmark of the human body pose; (2) Partition the body sub-parts of the foreground person. We use the partition information to extract latent representations of different parts. And they are further fused in the generator by skip connections in its U-Net structure. }\label{Fig intro1}
\end{figure}
Driven by the task requirements, we focus on the human image generation by changing the appearance, pose and background. Some algorithms have achieved pose changing in human image generation \cite{Siarohin_2018_CVPR}, but the background is not taken into account, causing blurred and random background which harms the quality of the whole image. \cite{DBLPjournalscorrabs-1712-02621} introduces a two-staged disentangled architecture to extract the features of body appearance, pose and background respectively. However, it projects the images to latent space with a black box encoder, which is difficult to adapt to more general applications like further editing body sub-parts.

To alleviate this problem, the proposed method tries to partition the image in a two-tier manner. As shown in Fig.\ref{Fig intro1}, we consider the task of changing the foreground person in the target image $x_{tgt}$ to the one in source image $x_{src}$. In the first tier, our operation is in the image space. We partition the image into foreground and background, and extract the landmarks of the human body in both images. While in the second tier, our operation is in the latent space. Inspired by the U-net structure in \cite{isola2017image}, we use two encoders (denoted as $E_1$ and $E_2$ below) and one decoder to build our W-Net. We feed $E_1$ with the source image and its landmarks to extract source appearance feature.
Meanwhile, we feed $E_2$ with target image and also its landmarks as well as background to extract pose feature and background feature. Then we connect those features in the decoder. Specifically, we concatenate target features to the output of decoder layers using the same skip connection like the U-net. In order to handle the pose variation problem between the source and target images, we introduce an affine transformation to the feature maps generated by $E_1$ so that the source appearance can better fit in the target image. Moreover, we partition both source and target foreground human body into 10 sub-parts. Using the one to one correspondence between them, the affine transformation can be learned.

In addition, we use two adversarial losses to constraint the human pose and appearance respectively. We also explore the combination of adversarial loss with the pixel-to-pixel loss to force the generated image being more realistic. The results show significant improvements on several datasets quantitatively and qualitatively.

Our contributions can be summarized as follows:
\begin{itemize}
  %\item We proposed a two-tier partition method. In tier-I, the method partition and project the foreground and the background into embedding space separately with a pair of encoders.
  %\item In tier-II, we partition the foreground into several body parts and transform them through the skip connections to the background embedding space.
  %\item To tolerate the misalignments caused by the torsion of the non-rigid transforms of human body, we implement the methods with GAN and perceptual loss. The experiments show great progress in changing background and partition-controlled human image generation.
  \item We propose a novel generation architecture PCGAN, in which we design a W-Net to fuse foreground and background features.
  \item We propose a new method to separate background and human foreground suitable for the human generation task.
  \item Our experiments show that our method can generate realistic human images but also produce the human pose and background as we want.
\end{itemize}
\section{Related Works}

\subsection{Image-to-image generation}

Most recently, image generation is mainly solved by Variational Auto-encoder (VAE) \cite{kingma2013auto} and GAN \cite{NIPS2014_5423}. VAE is designed on the probabilistic graphical model whose objective is maximizing the loglikelihood of two distributions. It first introduced the autoencoder (AE) \cite{HintonSalakhutdinov2006b} to generate images from noise distribution. While the original GAN tries to project Gaussian noise to the distribution of real images by a generator. The reason for its success lies in the proposed adversarial loss which distinguishes the generated data and real data using a discriminator.

The generator of GAN is built up with convolutional neural networks such as AE \cite{pathakCVPR16context,wang2016generative,salimans2016improved}, U-net structure \cite{isola2017image,RFB15a,DBLPjournalscorrYiZTG17}, and ResNet \cite{CycleGAN2017}. Specifically, the U-net consists of an encoder and a decoder, which are a set of convolutional and de-convolutional blocks respectively. The skip connections between encoder layers and decoder layers help to remain abundant levels information of the input images. In this paper, we also leverage the skip connection to combine partitioned features in the decoder to modify the output images.

Most of the image-to-image generation methods focus on changing appearance, such as, style transfer \cite{CycleGAN2017}, super-resolution \cite{DBLPjournalscorrLedigTHCATTWS16}, colorization \cite{DBLPjournalscorrZhangIE16,DBLPjournalscorrSangkloyLFYH16}, while they always lack prior knowledge for spatial deformations in their models. CycleGAN \cite{CycleGAN2017} does an excellent job on unpaired image transfer by introducing a reconstruction loss. However, they do not have the constraint on the geometric structure of objects in each image. Thus they show little structure difference between inputs and outputs. To alleviate the problem, some researchers propose conditional GAN to constraint the generator and discriminator with supervised knowledge and achieve better results \cite{isola2017image}. Since then, researchers can separate and edit different kinds of attributes in GAN based architectures. AttGAN \cite{he2017arbitrary} edits the face images by learning the latent representation of the attributes. However, there are few spatial deformations shown in the face images. StarGAN \cite{choi2017stargan} trains only one generator for several attributes on the face generation task, which simplifies the model of multi-domain image generation.

Compared with face editing, the human pose image generation is more complicated as the non-rigid structures cause more deformations and occlusions.
%\cite{reed2016generating} show how to generate images conditioned on part key points and segmentation masks.
\cite{ma2017pose} proposes a more general approach allowing arbitrary pose synthesis, which is conditioned on the image and pose key points. \cite{DBLPjournalscorrabs-1712-02621} integrates the appearance and pose of human body and background in latent space and generate the images from noise distribution. \cite{Siarohin_2018_CVPR} is closest to the proposed method, which modifies the foreground human pose by a deformable skip connection. But the background of the target image is not taken into consideration.

\subsection{Pose Estimation}

Image generation methods can be conditioned on a variety of side information such as sketch, text, landmarks, etc. The landmarks are widely used in face and human pose image generation. For example, in face image generation, the GAGAN \cite{DBLPjournalscorrabs-1712-00684} incorporates geometric information from the statistical model of facial shapes. The Super-FAN \cite{bulat2017super} proposes an end-to-end system to solve the landmark detection and super-resolution of human face images simultaneously. At meanwhile, a wide range of methods on human image generation is based on the human pose structure \cite{lassner2017generative,ma2017pose,Siarohin_2018_CVPR,DBLPjournalscorrabs-1712-02621}. But the pose landmarks usually cost expensive human annotations. The most recent researches have achieved real-time pose inferring in multiple people image. As the benchmarks and protocols in \cite{ma2017pose}, we obtain the human body pose by a state-of-the-art pose estimator \cite{cao2016realtime}. This may cause missing points in the training and testing input which is tolerable in the generative models.

\subsection{Segmentation}
The instance segmentation annotations are also widely used in the generation tasks. \cite{lassner2017generative} manipulates human images by changing the clothes in a given pose. But the model is costly in segmentation annotation and based on a complex 3D pose representation. Segmentation is a basic method in the computer vision field and is now developed rapidly in a short period of time. Driven by powerful baseline methods \cite{girshickICCV15fastrcnn,ren2015faster}, the segmentation task now can be solved in satisfactory time complexity and accuracy. We apply the state-of-the-art instance segmentation method Mask-RCNN to obtain the segmentation of the foreground against the background. The segmentation results serve as a supplementation of the pose estimation and help the partition method achieving more accurate prior knowledge.

\begin{figure}
  \centering
  \includegraphics[width=\linewidth]{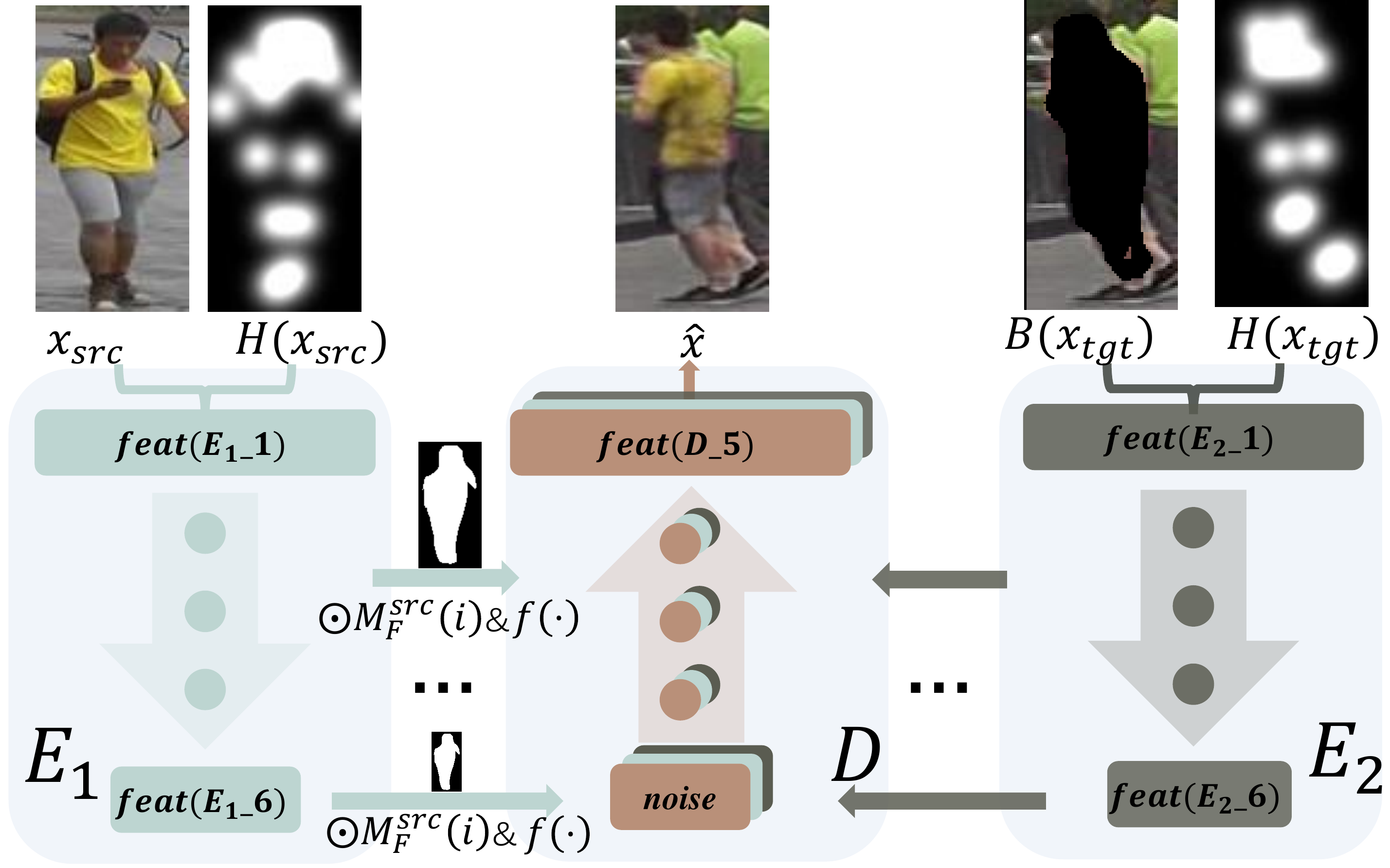}
  \caption{The generator (W-Net). Taken a pair of input $x_{src}$ and $x_{tgt}$, we first organize the input as the concatenation of images and heat maps (Eq.~\ref{Eq heat maps}). Note that for the target image $x_{tgt}$, we first split out the background using the mask extracted by Mask RCNN. To transmit the low-level information from both branches of the W-Net, we add skip connections from layers of $E_1, E_2$ to $D$. In order to adapt the feature maps to the decoder, we first add some transformations on them. $f(\cdot)$ is the affine transformation of each subpart. $\odot M_R^{src}(i)$ splits out the $i$-th body sub-part of $x_{src}$. Skip connections only employed in the first 4 layers.}\label{Fig G}
\end{figure}
\section{Partition-Controlled Image Generation}
In this section, we describe the proposed method of the partition-controlled generation networks.

Our goal is to manipulate the human image with the target pose and background. To that end, we propose the two-tier partition-controlled method. In tier-I, we partition the image into foreground and background, and extract the landmarks of the human body in both images in the image space. Considering the complexity of the non-rigid transformations of the foreground human body, we design the tier-II to partition and reconstruct the body parts in the latent space. In particular, by adopting the HPE and Mask-RCNN, we extract the pose landmarks and then decompose the human body into 10 parts with coarse masks. Then we use the affine transformation to project the features of 10 body parts to fit the background according to the target pose.

\subsection{Tier-I: Pose Estimation and Background Partition }
We first introduce some notations.
In short, we aim at generating images $\hat{x}$ which naturally change the foreground person (appearance and pose) to fit the background of another image.
Hence, different from the Deformable GAN \cite{Siarohin_2018_CVPR}, the generated image $\hat{x}$ is both conditioned on the pose $P(x_{tgt})$ and background $B(x_{tgt})$.
The model is trained on paired image datasets $\mathcal{X} = \{(x_{src}^{\{i\}}, x_{tgt}^{\{i\}})\}_{i = 1,\cdots, N}$, which contains the same person in different pose and background.
In the testing phase, we take two images $x_{src}$, and $x_{tgt}$ of different persons as input, where $x_{src}$ provides the person appearance and $x_{tgt}$ provides the pose and background.

To condition the generation model with the pose of the target image, we represent the landmarks as heat maps matrices.
Following the settings in \cite{Siarohin_2018_CVPR}, for each landmark $\mathbf{p}_j$, we represent the heat map matrix as a blurry area:
\begin{equation}\label{Eq heat maps}
  H_j(\mathbf{p}) = \exp\left(-\frac{\|\mathbf{p}-\mathbf{p}_j\|}{\sigma^2}\right),
\end{equation}
where $\sigma=6$, $\mathbf{p}$ runs through all the pixels in the heat map. Then we concatenate the $k$ heat maps of image $x$ to form the heat map tensor $H(x) = [H_1(\mathbf{p}), \cdots, H_k(\mathbf{p})], k = 18$.

To express the pose of a person in $x_{src}$, we extract $k$ landmarks using the Human Pose Estimator (HPE) $P(x) = (\mathbf{p}_1,\cdots,\mathbf{p}_k)$ as \cite{ma2017pose} did.
For fair comparison, we extract the same 18 landmarks as \cite{ma2017pose}.
In order to keep the background of $x_{tgt}$ in $\hat{x}$, we employ the masks $M(x_{tgt})$ to split out the backgrounds $B(x_{tgt}) = x_{tgt} \odot (\mathbf{1}-M(x_{tgt}))$.
In the meantime, to better fit the appearance of the foreground person to the background, we also split out the foreground person using the extracted mask of the source images $F(x_{tgt}) = M(x_{tgt}) \odot x_{tgt}$.
The masks of both the source image $M(x_{src})$ and the target image $M(x_{tgt})$ are extracted using the Mask-RCNN \cite{matterport_maskrcnn_2017} trained on COCO2014, with two classes person and background.
Both the landmarks and masks are used in the testing and training phase. Note that we do not use the ground-truth annotations for training.
Because the landmarks and masks are generated using HPE and Mask-RCNN, there are some missing points and parts (usually part of the foot, hand and head).
Also note that both the landmarks and masks can be done before training and testing.

%Each branch of the W-Net can be regarded as a U-net.

\subsection{Tier-II: Human Body Transformation}

The human body in the source image needs a more detailed transformation to fit the pose and background in the target image.
There are some algorithms which show competitive results on image translation tasks, but most of them focus on changing the appearances, such as making up and dressing.
In this paper, we change not only the color and style of the clothes, but also the body shape.
For example, we suppose to change a thin person in $x_{src}$ to the background in $x_{tgt}$, rather than change the appearance of the face and clothes to the $x_{tgt}$.
Similarly as \cite{Siarohin_2018_CVPR}, we decompose the human body to 10 sub-parts $R_i, i\in\{1,\cdots,10\}$: the torso, the head, the left/right upper/lower arm and the left/right upper/lower leg.
Here we made some adaption to the settings to take the body shape into consideration.
We define the body shape index according to the torso:
\begin{equation}\label{Eq body-shape-index}
  D_s = \frac{\|\mathbf{p}_{rh}-\mathbf{p}_{rs}\|_2+\|\mathbf{p}_{lh}-\mathbf{p}_{ls}\|_2}{2},
\end{equation}
where $\mathbf{p}_{ls}, \mathbf{p}_{rs}, \mathbf{p}_{lh}, \mathbf{p}_{rh}$ denotes left/right shoulder and left/right hip separately.
The head region is a square centered at landmarks left/right eye, left/right ear, nose with side length $0.8D_s$.
%The body region is a square centered at the middle of left/right shoulder and left/right hip with a side length $1.2D_s$ (a little larger than $D_h$ to take in the protruding parts).
The arms and legs contain 2 corresponding landmarks. We set these regions as rectangles with the width equal to $0.3D_s$.
%The legs and arms are rotated rectangles according to the 2 corresponding landmarks.

With the algorithms above, we obtain 10 rough body parts according to the pose of the source and target images.
However, they are still not exact enough to pick out the foreground from the background.
Some regions may not even exist because of the absence of the landmarks.
Hence, we extract the human body mask using Mask-RCNN and separate it into 10 body parts.
Take the source image for example.
First, we compute binary masks $M_R^{src}(i), i\in\{1,\cdots,10\}$ for each body part, which is zero everywhere except the points in the $i$-th sub-parts (defined above).
$M_F^{src}$ is obtained from Mask-RCNN, which is zero for the background and one for the person regions.
Then we calculate the Hadamard product of the body mask $M_F^{src}$ and the region masks $M_R^{src}(i)$ to gain the accurate mask:
\begin{equation}\label{Eq region-masks}
  M^{src}_F(i) = M_F^{src}\odot M_R^{src}(i).
\end{equation}
Considering the occlusion of the body parts, we define the torso as the rest of $M_F^{src}$ after removing the other sub-parts: $M_F^{src}(1) = M_F^{src}\backslash\{\cup_{2\cdots 10} M_F^{src}(i)\}$.
%By combining the results of HPE with Mask-RCNN, we correct the body masks and the body shape for the further processes.

\begin{figure}
  \centering
  \includegraphics[width=\linewidth]{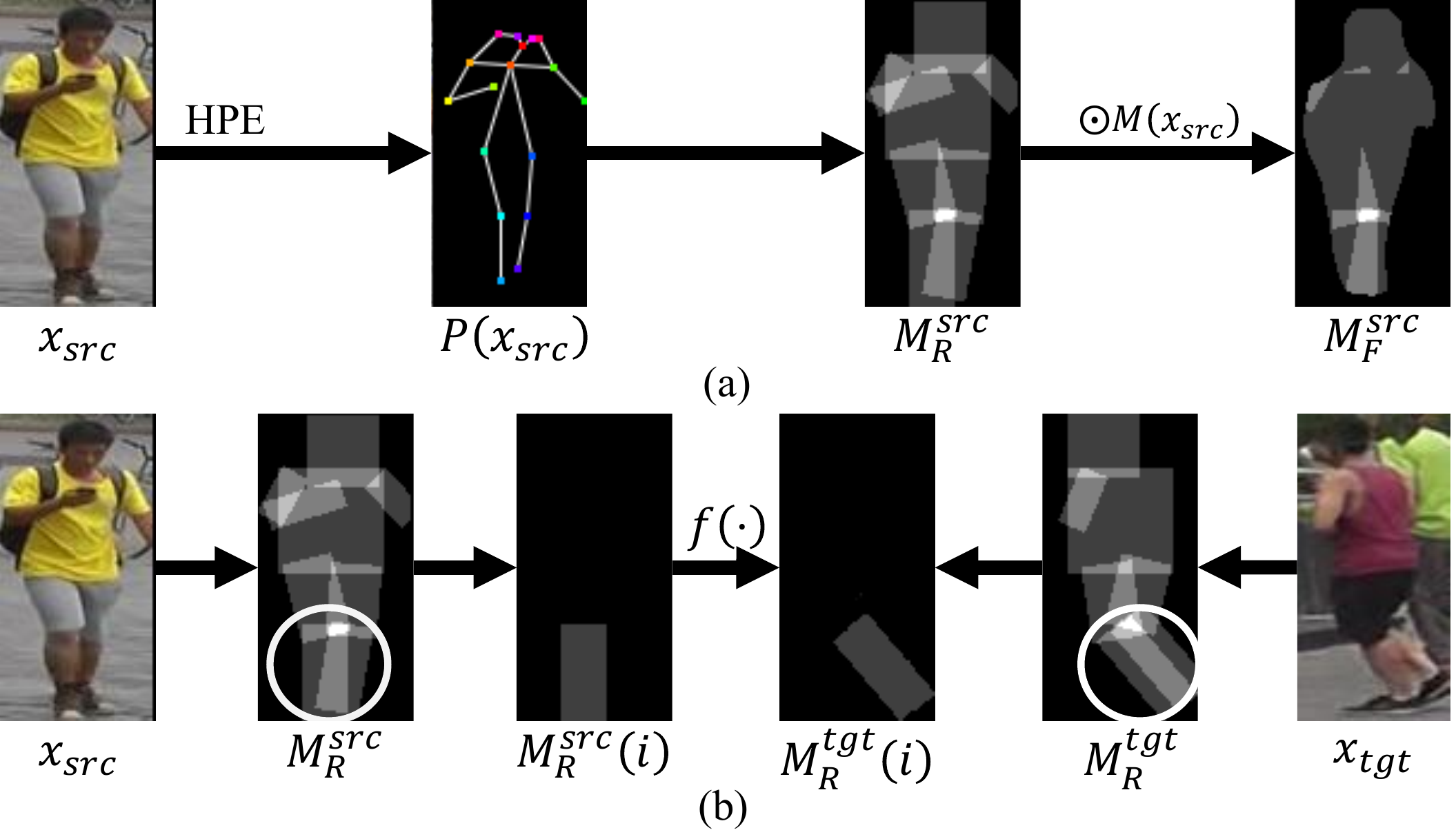}
  \caption{The first row shows the process of obtaining the masks of sub-parts. The second row shows the definition of the affine transformation on sub-parts.}\label{Fig mask}
\end{figure}

We define a set of affine transformations $f(\cdot)$ on sub-parts $M_R^{src}(i)$ to control the generation of the human body.
For example in Fig.~\ref{Fig mask}, for each body region, we transfer the region mask $M_R(x_{src})$ to $M_R'(x_{tgt})$ using $f(\cdot)$.
Note that we do not use the accurate mask $M^{src}_F$ to define the transformation, because of the difference of the body shape.
Moreover, the mask for $x_{tgt}$ uses the body shape index of $x_{src}$ in order to remain the shape of $x_{src}$ in the target image.
The parameters of the affine transformation are learned by minimizing a least square error.
Note that the affine transformation is adopted in the skip connections of the middle layers of the generator rather than the original images (see below).

\begin{figure*}
  \centering
  \includegraphics[width=\linewidth]{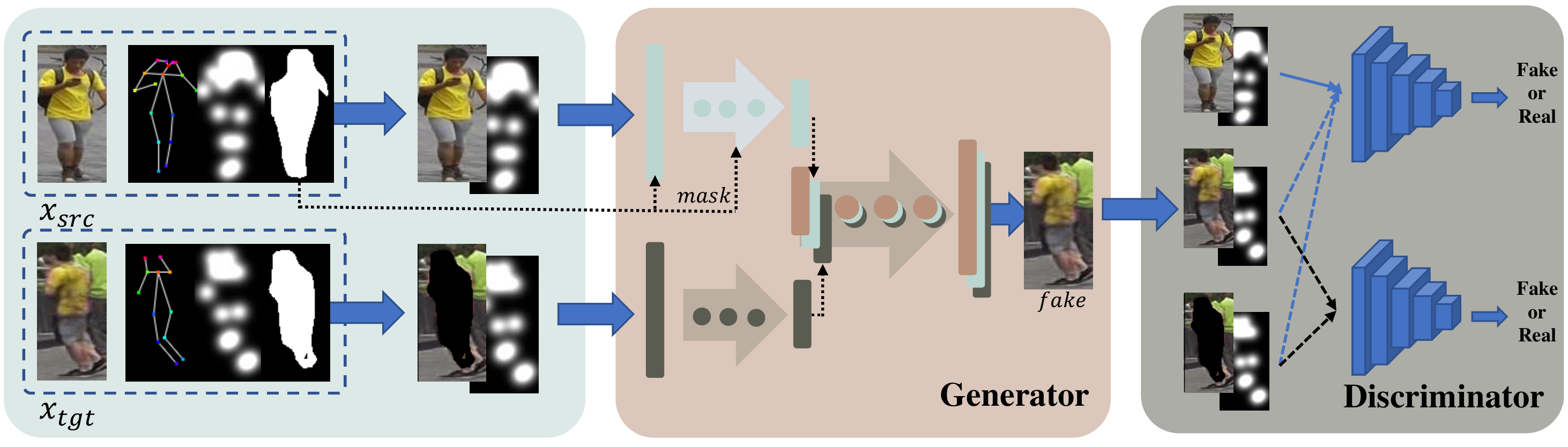}
  \caption{The whole PCGAN architecture. With the given input images $x_{src}$ and $x_{tgt}$, we first compute the poses $P(x_{src}),P(x_{tgt})$ and masks $M(x_{src}), M(x_{tgt})$. The heat maps $H(x_{src}),H(x_{tgt})$ are then calculated by pose. The generator is fed with $(x_{src},H(x_{src}))$ and $(x_{tgt}\odot M(x_{tgt}),H(x_{tgt}))$ with the masks conditioned on the skip connections on layers of `W-Net'.  Discriminator $D_1$ forces the generated images to have the same pose and appearance as the target images by conditioning on the ground-truth $(x_{tgt},H(x_{tgt}))$. $D_2$ only distinguishes the real and fake pairs. Both of them take $(x_{src}, H(x_{src}))$ as true and $(\hat{x},H(x_{tgt}))$ as false.}\label{Fig GAN}
\end{figure*}

\subsection{Network Architecture}
%Our first architecture aims to encode and decode the images by partitioning the images into foreground and background.
Inspired by \cite{NIPS2016_6544}, we propose a two-branched reconstruction architecture for the human body and the background separately, as shown in Fig.~\ref{Fig intro1}.
The generator $G$ takes as input 4 tensors: $(x_{src}, H(x_{src})$, $x_{tgt}$, $M(x_{tgt})\odot H(x_{tgt}))$.
Note that all these tensors can be pre-calculated before training and testing.

\subsubsection{W-Net}
In order to condition the generator with two streams of information, we propose a generator $G$ consists of two encoders and one decoder with skip connections (Fig.~\ref{Fig G}).
We name it by W-Net as each branch of it can be regarded as a U-net.
Encoder $E_1$ aims to extract the latent representations of the foreground appearance and pose $(x_{src}, H(x_{src}))$.
Moreover, the encoder $E_2$ is fed with the background appearance and pose tensor $(x_{tgt}\odot M(x_{tgt}), H(x_{tgt}))$.
These tensors are concatenated before passing into the encoders.
While both the two encoders aim at projecting the appearance and pose to latent space, we do not share the weights of the two encoders.
Because the appearance of the target background and the source foreground belongs to different categories, and also because the displacement of the landmarks is always too far regarding different poses.
By doing so, we guide the network to remain not only the appearance but also the spatial information in the feature maps.
Then, the decoder fuses the feature maps extracted from $E_1$ and $E_2$ by U-net structure with skip connections to join the foreground and the background together.

As described in the last section, we design the feature maps of the generator highly related to the location of the landmarks.
In order to partially control the generation of the human body of source images $x_{src}$, we apply affine transformations to the feature maps generated by $E_1$.
Specifically, we do the affine transformation on each sub-parts of the human body in the latent space. Then we add them together and skip connect them with the layers in the decoder (Fig.~\ref{Fig G}).

\subsubsection{Discriminator}
We design two discriminator networks.
$D_1$ is a fully-convolutional discriminator conditioned on the given ground-truth appearance and pose.
Specifically, we concatenate $(x_{src}, H(x_{src}), x_{tgt}, H(x_{tgt}))$ as positive examples and $(x_{src}, H(x_{src}), \hat{x}, H(x_{tgt}))$ as negative examples.
But the texture of persons on the two compared images is not aligned strictly landmark by landmark.
To alleviate the misalignment, we use another discriminator $D_2$ to distinguish fake images and real images.
It  takes as input an image $x_{src}$ or $x_{tgt}$.
Both the discriminators return scalar values in $[0,1]$ denote the probability that the input $\hat{x}$ is a real image.

\subsubsection{Objective functions}

We train the whole network with two standard GAN loss. The cGAN loss conditioned by the ground-truth is given by:
\begin{equation}\label{Eq cGAN-loss}
  L_{cGAN}(G,D_1) = E[\log D_1(x|y)] +  E[\log(1- D_1(\hat{x}|y))].
\end{equation}
where $x$ is the concatenation of real inputs $(x_{src}, H(x_{src}))$ and $\hat{x}$ is the output of the generator $(\hat{x}, H(x_{tgt}))$.
The condition $y$ denotes $(x_{tgt}, H(x_{tgt}))$.

The GAN loss distinguishing real and fake image domains is given by:
\begin{equation}\label{Eq GAN-loss}
  L_{GAN}(G,D_2) = E[\log D_2(x_{tgt})] +  E[\log(1- D_2(\hat{x}))].
\end{equation}

We also apply $L_1$ loss to the reconstructed image for the generator:
\begin{equation}\label{Eq L1-loss}
  L_1(\hat{x},x_{tgt}) = \|\hat{x}-x_{tgt}\|_1.
\end{equation}

%Moreover, we apply the perceptual loss to the generated images.
%To deal with the occlusion and miss-detection of landmark, we would like to use the perceptual loss to help the generated images complementing body parts and masks.
%The middle layers of a pre-trained neural networks can extract features representing high-level semantic information.
%We adopt the VGG-19 \cite{DBLPjournalscorrSimonyanZ14a} trained on ImageNet and extract the feature map from $conv_{4\_2}$ for feature loss:
%\begin{equation}\label{Eq Perceptual-loss}
%  L_P(\hat{x},x_{tgt}) = \|F_{4\_2}(\hat{x})-F_{4\_2}(x_{tgt})\|_2.
%\end{equation}

Combining Eq.~\ref{Eq cGAN-loss} - Eq.~\ref{Eq L1-loss}, the objective function is:
\begin{eqnarray}\label{Eq obj}
  L = \min_{G}\max_{D} L_{cGAN}(G,D_1) + \lambda_1L_{GAN}(G,D_2)&& \nonumber\\
   + \lambda_2L_1(\hat{x},x_{tgt}).&&
\end{eqnarray}

The parameters are set $\lambda_1 = 1, \lambda_2 = 0.01$ in the experiments.
%\begin{eqnarray}
%% \nonumber to remove numbering (before each equation)
%  L_{GAN}(G,D_1) = E[\log D(x_{src}, H(x_{src}), x_{tgt}, H(x_{tgt}))]&& \nonumber\\
%  +  E[\log(1- D(x_{src}, H(x_{src}), \hat{x}, H(x_{tgt})))] &&\nonumber\\
%   &&
%\end{eqnarray}

%%
\section{Experiments}
In this section, we introduce the experiment settings and the training details of the algorithm.
Both qualitative and quantitative results are demonstrated on the human image generation task.
The experiment results show that the proposed method outperforms the existing algorithms especially cases with complex geometry deformations.
Code can be found at \url{https://github.com/AlanIIE/PCGAN}.
% and additional experiments
\subsection{Training Procedure}

We train all the networks with mini-batch Adam optimizer (learning rate: 2e-4, $\beta_1$ = 0.5, $\beta_2$ = 0.999).
The generator is designed in U-net structure with two encoders and one decoder.
Our encoder is set as 7 blocks for the DeepFashion dataset and 6 blocks for the Market-1501 dataset.
Let \texttt{c3s1-k} denote a $3\times3$ Convolution-ReLU layer with k filters and stride 1.
\texttt{dk} denotes a $4\times4$ Convolution-InstanceNorm-ReLU layer with k filters and stride 2.
\texttt{uk} denotes a $4\times4$ fractional-strided-Convolution-InstanceNorm-ReLU layer with k filters and stride 1/2.

The encoder network with 6 blocks consists of:
\begin{verbatim}
c3s1-64,d128,d256,d512,d512,d512
\end{verbatim}
The encoder network with 7 blocks supplements the 6 blocks with an additional \texttt{d512} behind.

The corresponding decoder is:
\begin{verbatim}
u512,u512,u512,u256,u128,c3s1-3
\end{verbatim}
The first 3 layers use dropout at rate 50\% at training time.

The discriminator architecture is:
\begin{verbatim}
c4s2-64,d128,d256,d512,d1
\end{verbatim}
We replace the ReLU of the last layer with sigmoid.
The InstanceNorm layer in the last layer of both the encoder and the discriminator is removed.

The network is trained 90 epochs with 500 iterations.
In each iteration, we typically set the critic per iteration per generator iteration to 2.
The discriminator is trained 2 steps before the generator thus the discriminator can provide more reliable results \cite{DBLPjournalscorrYiZTG17}.
\subsection{Datasets and Metrics}

\begin{figure*}
  \centering
  \includegraphics[width=\linewidth]{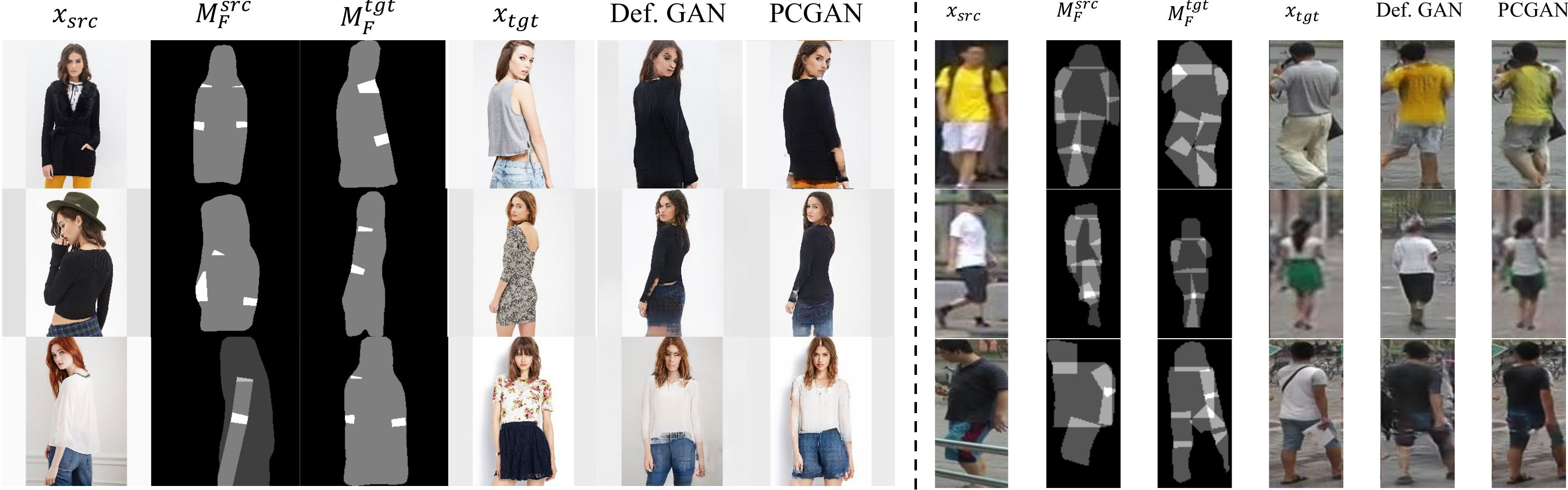}
  \caption{From left to right: qualitative results compared with Deformable GAN (Def. GAN) on DeepFashion and Market-1501.}\label{Fig comp_res2}
\end{figure*}

\begin{figure}[h]
  \centering
  \includegraphics[width=\linewidth]{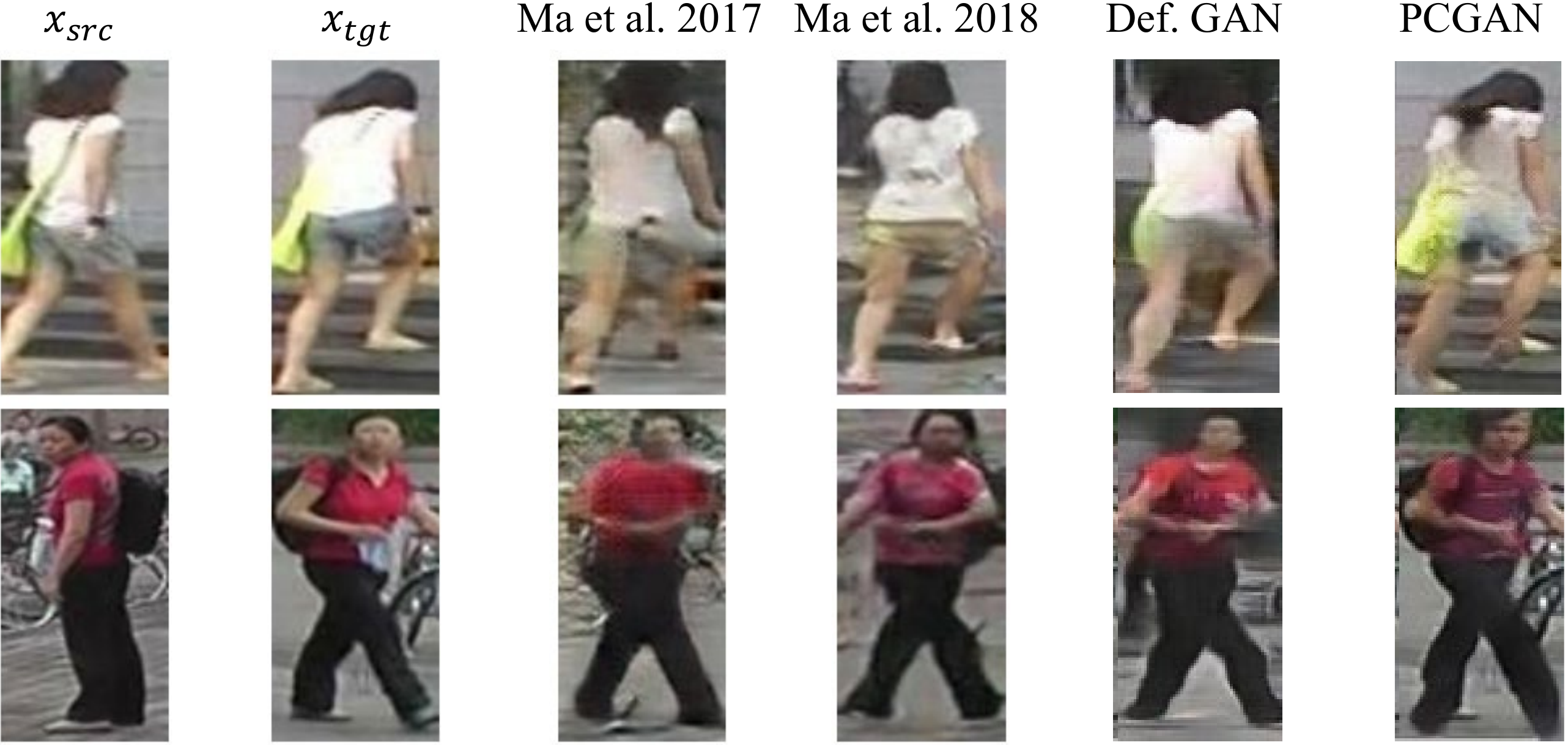}
  \caption{Qualitative results compared with \cite{ma2017pose}, \cite{DBLPjournalscorrabs-1712-02621} and Deformable GAN \cite{Siarohin_2018_CVPR}. The results of \cite{ma2017pose}, \cite{DBLPjournalscorrabs-1712-02621} are adopted from their paper. Other results are tested on the corresponding algorithm by their released checkpoints.}\label{Fig comp_res}
\end{figure}

\normalsize

We use the person re-identification dataset Market-1501 \cite{zheng2015scalable}, containing 32,668 images of 1,501 persons captured from 6 disjoint surveillance cameras. All images are resized to $128\times64$ pixels. The dataset is challenging because of the multiple appearance and pose of human, and the diversity of the illumination background and viewpoint.
We first run the HPE and Mask-RCNN to obtain the landmark and segmentation results.
Then we remove the images with no human body detected, and results in 232,696 training pairs and 10,560 testing pairs.
No common person occurs in both the training and testing sets.

We also conduct experiments with a high-resolution dataset DeepFashion (In-shop Clothes Retrieval Benchmark) \cite{liu2016deepfashion}, that composed of 52,712 in-shop clothes images and 200,000 cross-pose/scale pairs.
The image pairs are collected as two different poses/scales of the same persons wearing the same clothes.
We split the dataset to training/test following the settings in \cite{ma2017pose}.
After removing the images which cannot detect the human body by HPE and Mask-RCNN, we finally select 32,008 pairs for training and 7,662 pairs for testing.

The COCO2017 dataset \cite{lin2014microsoft} is a large-scale dataset for multiple computer vision tasks including segmentation. The images are annotated with instance segmentation labels.
According to the annotation, we pick out the images with person whose bounding box is larger than $128\times64$.
Then we cut the images out according to the bounding box and padding it to length-width ratio 2:1 with zeros and resize it to $128\times64$.
After removing the images which fail to detect human body by HPE, we obtain 47,153 images (there are some images obtained from one original image).
We random select 10,000 pairs only for testing.

The LIP \cite{liang2015human} is a dataset focusing on the semantic understanding of person.
We use the annotations of clothes as the mask of person and do the same process as above.
Finally, we obtain 40,462 images and randomly select 500 pairs for testing.

For quantitative study, we employ the inception score (IS) \cite{salimans2016improved} and the masked versions mask-IS \cite{ma2017pose}.
Moreover, the FID \cite{DBLP:conf/nips/HeuselRUNH17} is also introduced to capture the similarity of generated images to real ones.
We note that lower FID score is better and IS should be larger for better images.
Different from \cite{ma2017pose}, we do not use the SSIM because for this task, there is no ground-truth to calculate the similarity.
We also depict some qualitative results.

\begin{table}%[htbp]
\scriptsize
  \centering
  \caption{Quantitative comparison with the state-of-the-art.}
    \begin{tabular}{c|ccc|cc}
\hline
          & \multicolumn{3}{c|}{Market-1501} & \multicolumn{2}{c}{DeepFashion} \\
\hline
    Model                                  & IS ($\Uparrow$)   & mask-IS ($\Uparrow$) & FID ($\Downarrow$)   & IS ($\Uparrow$)   & FID ($\Downarrow$) \\
    \cite{ma2017pose}                      & 3.460             & 3.435                & -                    & 3.090             & - \\
    \cite{DBLPjournalscorrabs-1712-02621}  & 3.483             & 3.491                & -                    & 3.228             & - \\
    Def. GAN              & 3.185             & 3.502                & 45.958               & 3.439             & \textbf{19.200} \\
    PCGAN                                  & \textbf{3.657}    & \textbf{3.614}       & \textbf{20.355}      &\textbf{ 3.536}    & 29.684 \\
\hline
    Real-Data & 3.860  & 3.360  & 6.811 & 3.898 & 3.446  \\
\hline
    \end{tabular}%
  \label{tab overall}%
\end{table}%

\subsection{Image Manipulation}

\begin{figure*}
  \centering
  \includegraphics[width=\linewidth]{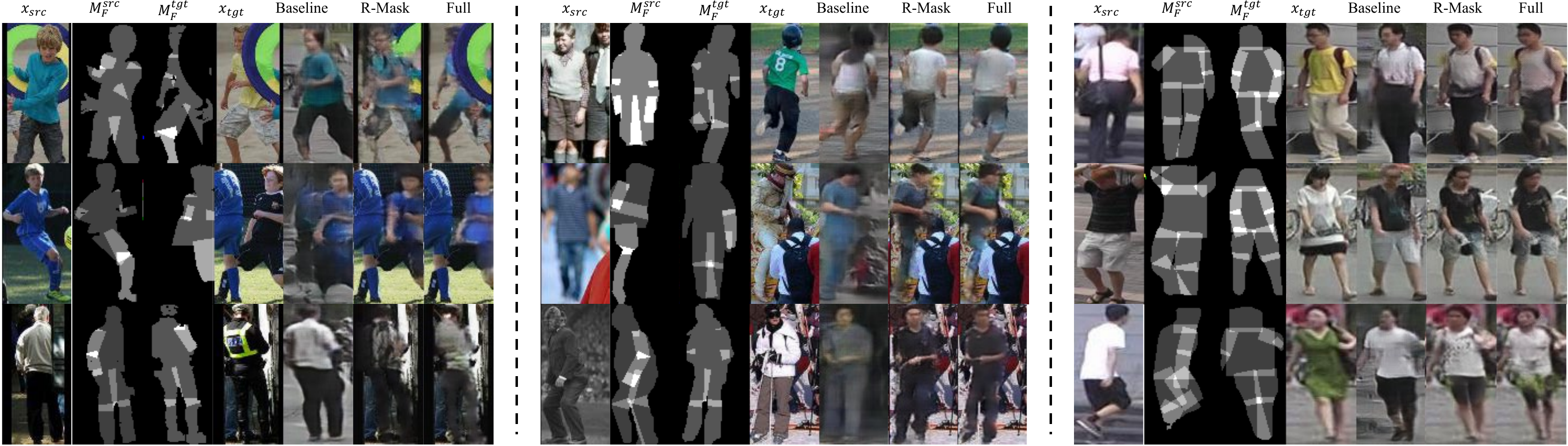}
  \caption{From left to right: sample results of the ablation study on LIP, COCO2017, Market-1501.}\label{Fig ablation}
\end{figure*}

%\begin{figure}
%  \centering
%  \includegraphics[width=\linewidth]{img/market.pdf}
%  \caption{market}\label{Fig market}
%\end{figure}
%
%\begin{figure}
%  \centering
%  \includegraphics[width=\linewidth]{img/fasion.pdf}
%  \caption{DeepFashion}\label{Fig fashion}
%\end{figure}

As described above, the partition-controlled algorithm decomposes the images to three factors: appearance, pose and background.
We train the network on the Market-1501 and DeepFashion dataset and test the algorithm by fitting the person in the source image to the background of the target image.
The tests on COCO2017 and LIP use the weights trained on the Market dataset.

As shown in Fig.~\ref{Fig comp_res2} and Fig.~\ref{Fig comp_res}, the proposed method generates more realistic images with background given by the target image.
The clear background is a progressive improvement from the baseline method, which has more realistic details and fewer artifacts.
In the meantime, our method can achieve competitive results in the DeepFashion dataset which has simple background.
This is mainly caused by the partition-controlled architecture which splits out the foreground and background of an image in the pixel level, and manipulating the pose by editing its latent representation.
The qualitative results agree with the quantitative results in Table.~\ref{tab overall}.
The proposed method performs better than the other methods in most cases.

\subsection{Ablation Study}

We also try ablation study on masks and loss functions.
We first list the methods to compare below:

\begin{itemize}
  \item Baseline: We use the Deformable GAN setting, which changes the pose of the human body in the source image without further changing the background.
  We implement the experiment with the released code and checkpoints.
  \item R-mask: We use the Region mask to process the feature maps in the skip connections of the generator, where other settings are the same as in the full-pipeline.
  \item Full: The proposed method using more accurate mask based on the Mask-RCNN (Market, Fashion), or ground-truth annotation (LIP, COCO2017).
\end{itemize}

% Table generated by Excel2LaTeX from sheet 'Sheet3'

\begin{figure}
  \centering
  \includegraphics[width=\linewidth]{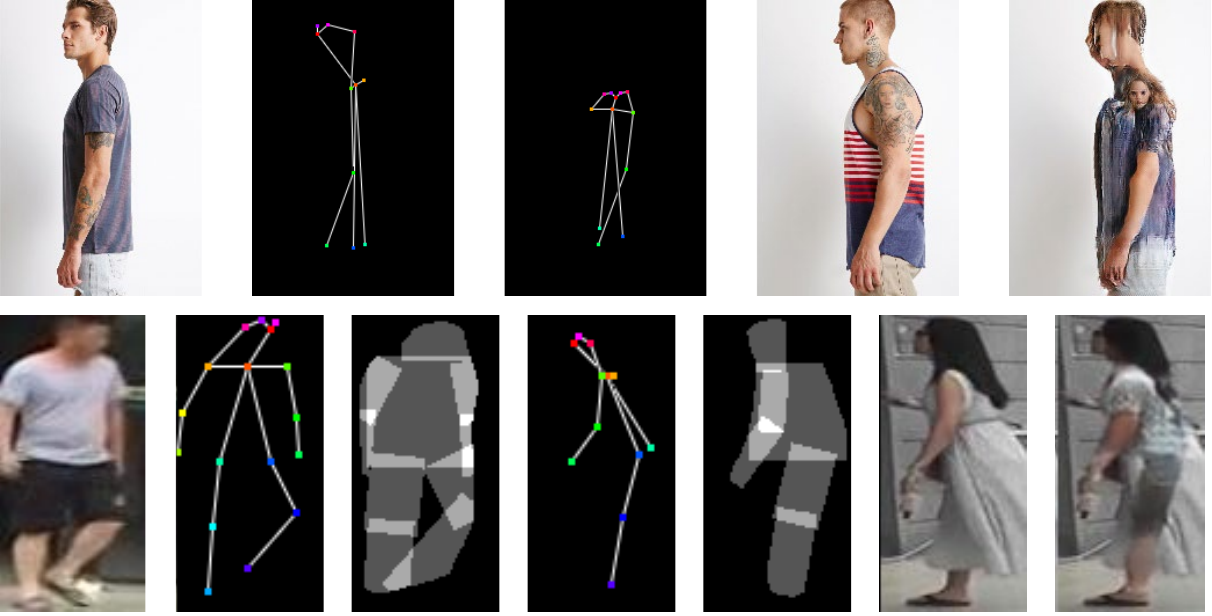}
  \caption{Several failure cases}\label{Fig: failure cases}
\end{figure}

As shown in Table.~\ref{tab ablation}, there are significant improvements from the baseline method to the R-Mask.
Full is slightly better than R-Mask in most datasets as the Mask-RCNN improves the accuracy of the masks.
We ascribe the lower score on the market dataset to the low resolution of the images, which causes bad segmentation results (missing head /foot /hand) and thus bad generation results.

We depict some qualitative results in Fig.~\ref{Fig ablation}. We can see that accurate segmentation mask can help the generation in LIP and COCO2017 with segmentation annotations.
For images with high resolution and clear background, the proposed method also achieves competitive results.

We have also categorized the failure cases in Fig.~\ref{Fig: failure cases}.
The first case is related to errors in HPE when given pose landmarks.
The model also fails when the region is so large that the mask cannot cover the expected body part.

%\begin{figure}
%  \centering
%  \includegraphics[width=\linewidth]{img/COCO.pdf}
%  \caption{COCO}\label{Fig COCO}
%\end{figure}
%
%\begin{figure}
%  \centering
%  \includegraphics[width=\linewidth]{img/LIP.pdf}
%  \caption{LIP}\label{Fig LIP}
%\end{figure}

\begin{table}%[htbp]
  \centering
  \caption{Quantitative ablation study (IS scores).}
    \begin{tabular}{ccccc}
\hline
    Model & LIP   & COCO2017 & Market & Fashion \\
\hline
    Baseline & 3.383  & 3.576  & 3.185  & 3.168  \\
    R-Mask & 4.794  & 5.237  & \textbf{3.821}  & 3.168  \\
    Full  & \textbf{4.957}  &\textbf{ 5.448}  & 3.657  & \textbf{3.536}  \\
\hline
    \end{tabular}%
  \label{tab ablation}%
\end{table}%

\section{Conclusion}

In this paper, we present a two-tier image generation method solving the human image manipulation problem.
Tier-I splits the foreground and background images in pixel level, while tier-II decomposes the body sub-parts in latent space.
The generator is designed as a W-Net which integrals the information form both encoders to one decoder by skip connections.
In this way, we not only emphasis details in human body, but also remain the background in the generated images.
Experiments show that the proposed method achieves excellent results in human image generation with the input source foreground and target background.
Beyond the testing on Market-1501 and DeepFashion following \cite{ma2017pose}, tests are also done on the COCO2017 and LIP.
However, image manipulating remains a hard problem, especially for the unpaired data.

\section{ Acknowledgments}
Supported by the National Key R\&D Program of China (Grant No. 2016YFC0801004). National Natural Science Foundation of China (No.U1736219，U1605252).

\bigskip
\bibliography{AAAI_bib}
\bibliographystyle{aaai}
\end{document}